\documentclass[runningheads,a4paper]{llncs}
\usepackage[T1]{fontenc}
\usepackage{graphicx}
\usepackage{amsmath}
\usepackage{amssymb}
\usepackage{booktabs}
\usepackage{url}
\usepackage{xcolor}
\usepackage{microtype}
\usepackage{marvosym}
\definecolor{linkblue}{RGB}{0,82,155}
\usepackage[colorlinks=true,citecolor=linkblue,linkcolor=linkblue,urlcolor=linkblue]{hyperref}
\usepackage{orcidlink}
\begin{document}
\title{CSG-Mamba: A Convolutional Scoring Gating Vision State Space Network for Endoscopic Polyp Segmentation}
\titlerunning{CSG-Mamba for Endoscopic Polyp Segmentation}
\author{Yuliang Wang\textsuperscript{*}\,\orcidlink{0009-0009-5357-8547} \and
Jiaqi Wu\textsuperscript{*}\,\orcidlink{0009-0009-9365-6896} \and
Jiaye Song\,\orcidlink{0009-0000-1837-4703} \and
Shuxia Ren\textsuperscript{(\Letter)}}
\authorrunning{Y. Wang et al.}
\institute{School of Advanced Interdisciplinary Studies, Tiangong University, Tianjin, China\\
\email{t\_rsx@126.com}}
\maketitle
\begin{NoHyper}
\begingroup
\renewcommand{\thefootnote}{*}
\footnotetext{These authors contributed equally to this work.}
\endgroup
\end{NoHyper}
\begin{abstract}
Accurate polyp segmentation is critical for computer-aided colonoscopy, yet endoscopic images often contain low-contrast boundaries, mucosal texture interference, specular highlights, and device-dependent appearance shifts. Vision State Space Models (SSMs) provide efficient long-range modeling with linear complexity, but existing Vision Mamba segmentation models typically convert 2D features into 1D scanning sequences, which may weaken local geometric continuity and over-smooth irregular contours. We propose CSG-Mamba, a convolutional scoring gating Vision State Space network for endoscopic polyp segmentation. Built on a VM-UNet-style asymmetric U-shaped encoder-decoder, CSG-Mamba inserts a Convolutional Scoring Gating (CSG) module at the semantically rich bottleneck. CSG generates a local spatial score map through pointwise and large-kernel depthwise convolutions and recalibrates state-space features by multiplicative gating. Experiments with three random seeds show that CSG-Mamba achieves 0.9220 Dice and 15.87 HD95 on Kvasir-SEG, and 0.7418 Dice and 0.6570 mIoU on CVC-ColonDB, outperforming the baselines on most overlap and recall metrics while maintaining competitive boundary accuracy. Code is available at \url{https://github.com/GGMANUTD/CSG-Mamba}.

\end{abstract}
\section{Introduction}

Polyp segmentation is a fundamental task in computer-aided endoscopic analysis. Real colonoscopy images are affected by low contrast, mucosal texture interference, specular highlights, motion blur, and substantial variations in lesion scale and appearance. A practical model must therefore achieve high in-domain overlap while maintaining robustness on external datasets collected with different devices and acquisition protocols.

Fully convolutional networks established end-to-end dense prediction~\cite{long2015fcn}. U-Net~\cite{ronneberger2015unet}, UNet++~\cite{zhou2018unetpp}, and residual U-shaped variants~\cite{he2016resnet,zhang2018resunet} provide strong local inductive bias, while PraNet~\cite{fan2020pranet} improves polyp segmentation through reverse attention. However, pure convolution is limited in modeling long-range dependencies. Transformer-based methods such as TransUNet~\cite{chen2021transunet} and PVT-style dense prediction backbones~\cite{wang2021pvt} alleviate this limitation through self-attention or hierarchical token representations, but their memory and computational costs remain high for dense prediction. State Space Models (SSMs), from S4~\cite{gu2022s4} to Mamba~\cite{gu2023mamba}, provide linear-time sequence modeling. Vision Mamba~\cite{zhu2024vim}, VMamba~\cite{liu2024vmamba}, VM-UNet~\cite{ruan2024vmunet}, VM-UNetV2~\cite{zhang2024vmunetv2}, and SegMamba~\cite{xing2024segmamba} extend this modeling paradigm to visual and medical segmentation tasks.

The key limitation is that visual Mamba models rely on sequential scanning, whereas low-contrast segmentation depends strongly on 2D spatial continuity. SS2D unfolds 2D features into 1D sequences along multiple directions, expanding contextual coverage while potentially weakening local adjacency around fine boundaries. This mismatch may lead to over-smoothing, under-segmentation, and boundary displacement.

We propose CSG-Mamba to complement global state-space modeling with local convolutional scoring. A lightweight Convolutional Scoring Gating (CSG) module is inserted at the bottleneck, where semantic information is concentrated and the spatial resolution is compact. CSG generates a spatial score map through pointwise and large-kernel depthwise convolutions and recalibrates features by multiplicative gating. To isolate the effect of the proposed module, the Mamba-based models are trained under the same zero-inference-cost stabilization protocol, including deep supervision, EMA, and DropPath.

The main contributions of this work are summarized as follows.
\begin{itemize}
\item We propose CSG-Mamba for endoscopic polyp segmentation. The method introduces convolutional scoring gating into the bottleneck of a U-shaped Vision State Space network, using lightweight local convolutional priors to compensate for the loss of 2D spatial continuity caused by SS2D sequentialization.
\item We design the CSG module by combining large-kernel convolutional scoring with multiplicative gating, allowing local boundary-aware responses to regulate global state-space features without introducing matrix attention.
\item We adopt a consistent training protocol for the Mamba-based comparisons, using deep supervision, EMA, and stochastic depth to reduce optimization variance without adding inference overhead.
\item We conduct experiments with three random seeds on Kvasir-SEG, CVC-ClinicDB, and CVC-ColonDB, showing favorable parameter efficiency, boundary accuracy, and cross-domain generalization under this matched protocol.
\end{itemize}

\section{Related Work}
\subsection{Polyp Image Segmentation}

FCN~\cite{long2015fcn} first showed that classification networks can be converted into dense predictors, and U-Net~\cite{ronneberger2015unet} established the dominant encoder-decoder framework for medical image segmentation. UNet++~\cite{zhou2018unetpp} narrows the semantic gap between encoder and decoder features through nested skip pathways and deep supervision. Residual learning~\cite{he2016resnet} further improves optimization and leads to variants such as ResUNet~\cite{zhang2018resunet}. For polyp segmentation, PraNet~\cite{fan2020pranet} uses a parallel partial decoder and reverse attention to mine boundary regions. These CNN-based methods are stable and efficient, but their local receptive fields limit global shape modeling. DeepLabv3+~\cite{chen2018deeplabv3plus} and Transformer models, including TransUNet~\cite{chen2021transunet} and PVT~\cite{wang2021pvt}, introduce broader context through atrous convolution, self-attention, or hierarchical tokens. Nevertheless, modeling long-range dependencies remains expensive for high-resolution dense prediction. This work instead focuses on Vision State Space Models and improves their local boundary representation.

\subsection{Vision State Space Models}

S4~\cite{gu2022s4} enables deep long-sequence learning with structured state-space parameterization, and Mamba~\cite{gu2023mamba} introduces input-dependent selective modeling while preserving linear complexity. Vision Mamba~\cite{zhu2024vim} and VMamba~\cite{liu2024vmamba} adapt selective state-space modeling to visual inputs, with VMamba using SS2D to scan 2D feature maps. In medical segmentation, VM-UNet~\cite{ruan2024vmunet}, VM-UNetV2~\cite{zhang2024vmunetv2}, U-Mamba~\cite{ma2024umamba}, and SegMamba~\cite{xing2024segmamba} show that state-space blocks can provide efficient contextual modeling. However, SS2D still converts local 2D neighborhoods into scan-order relationships. For low-contrast pixel-level segmentation, this conversion can weaken contour integrity. We therefore introduce local convolutional priors at the module level so that Mamba-style global modeling and CNN-style boundary bias can complement each other.

\subsection{Attention, Gating, and Deep Supervision}

Attention U-Net~\cite{schlemper2019attention} and CBAM~\cite{woo2018cbam} show that explicit feature selection can suppress background interference. Large-kernel and depthwise convolution can also expand local receptive fields at low cost~\cite{chen2018deeplabv3plus,liu2022convnext}. Inspired by these ideas, CSG generates spatial scores using pointwise and depthwise convolutions and applies multiplicative feature gating without matrix attention. Deep supervision~\cite{lee2015deeply}, used in UNet++~\cite{zhou2018unetpp} and medical frameworks such as nnU-Net~\cite{isensee2021nnunet}, is combined with EMA in our training protocol to stabilize optimization on small datasets. The auxiliary heads are removed during inference.

\section{Method}
\subsection{Overall Architecture}

\begin{figure}[t]
\centering
\includegraphics[width=.98\textwidth]{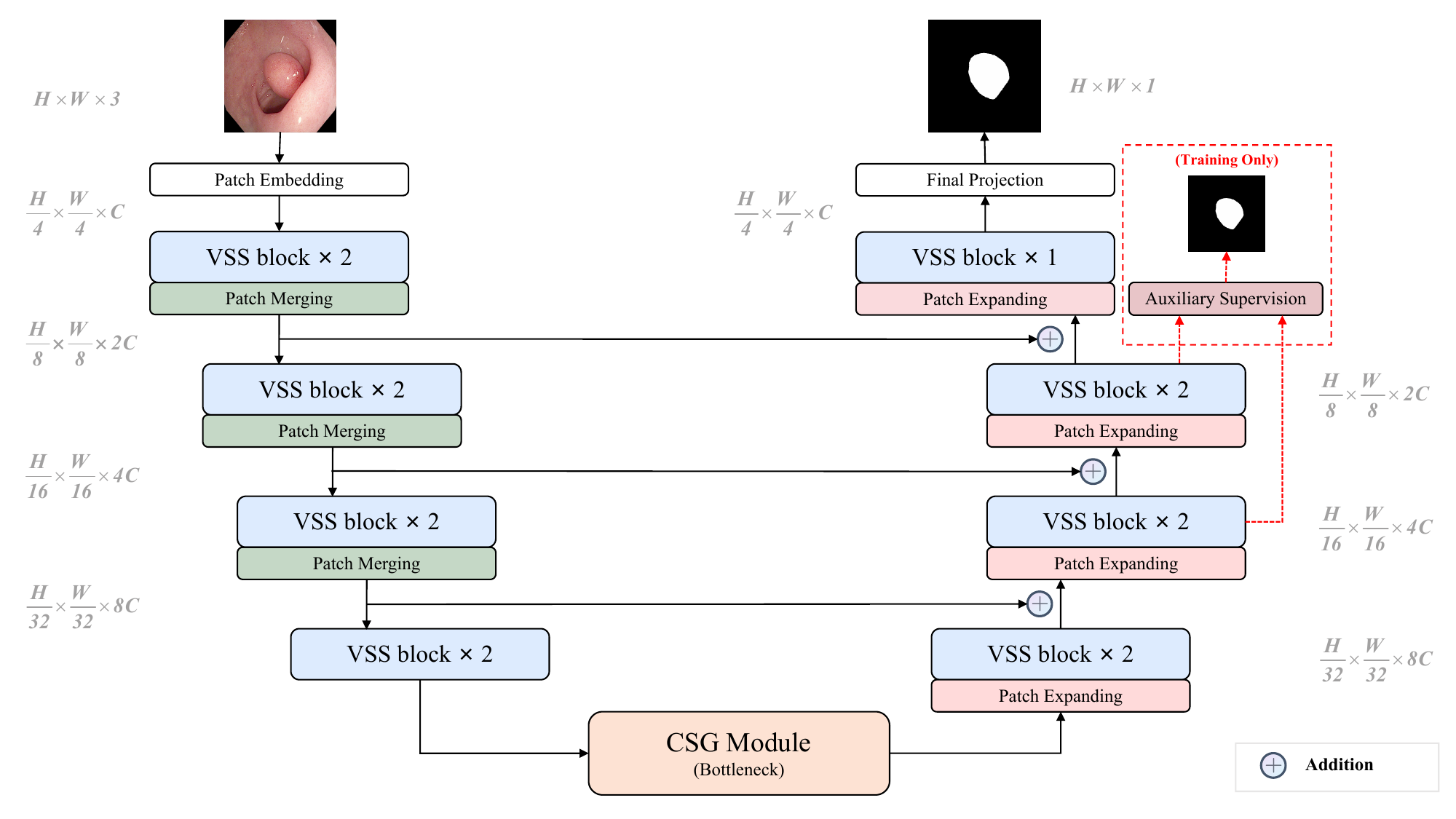}
\caption{Overall architecture of CSG-Mamba. The CSG module is inserted at the bottleneck, and auxiliary supervision is used only during training.}
\label{fig:overall}
\end{figure}

Fig.~\ref{fig:overall} illustrates CSG-Mamba. Given an image $\mathbf{X}\in\mathbb{R}^{H\times W\times 3}$, Patch Embedding projects it to $\mathbf{F}_0\in\mathbb{R}^{\frac{H}{4}\times\frac{W}{4}\times C}$. The encoder stacks VSS Blocks and uses Patch Merging to reduce the resolution and increase the channel dimension. At the deepest stage, a CSG module recalibrates the bottleneck feature, where semantic information is concentrated and the convolutional overhead is low. The decoder uses Patch Expanding and skip connections to recover spatial resolution, followed by a $1\times1$ prediction head.

\begin{figure}[t]
\centering
\includegraphics[width=.88\textwidth]{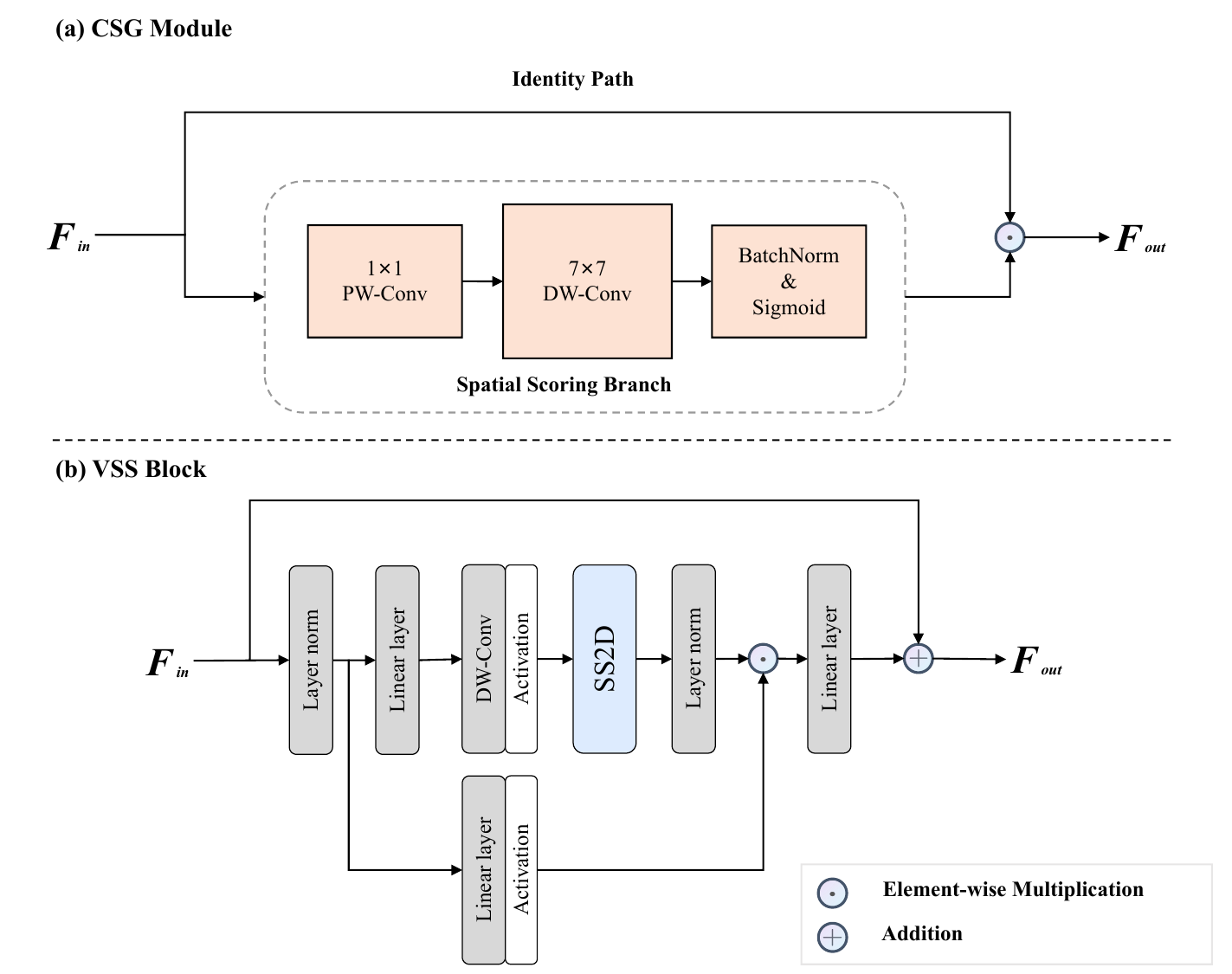}
\caption{Detailed structures of the proposed components: (a) CSG module; (b) VSS Block.}
\label{fig:csg_vss}
\end{figure}

The implementation follows VMamba-Tiny. Encoder channels are 96, 192, 384, and 768 with depths $[2, 2, 2, 2]$, and decoder depths are $[2, 2, 2, 1]$. As detailed in Fig.~\ref{fig:csg_vss} (b), each VSS Block sets the state dimension to 16, the local convolution kernel size to 3, the expansion ratio to 2, and the time-step projection rank to $\lceil d_{\mathrm{model}}/16\rceil$. PatchEmbed2D is implemented as a $4\times4$ stride-4 convolution. PatchMerging2D projects $4C$ channels to $2C$, while PatchExpand2D doubles the spatial resolution and halves the channel dimension. The final expansion upsamples the feature map by a factor of 4 to produce the segmentation mask.

\subsection{Visual State Space Block and SS2D}

A continuous state-space model can be formulated as
\begin{equation}
\frac{d\mathbf{h}(t)}{dt}=\mathbf{A}\mathbf{h}(t)+\mathbf{B}x(t),\quad
y(t)=\mathbf{C}\mathbf{h}(t),
\label{eq:ssm_continuous}
\end{equation}
where $x(t)$ denotes the input, $\mathbf{h}(t)$ denotes the hidden state, $y(t)$ denotes the output, and $\mathbf{A}$, $\mathbf{B}$, and $\mathbf{C}$ are learnable or input-dependent parameters. After discretization, the model can be written as
\begin{equation}
\mathbf{h}_k=\bar{\mathbf{A}}\mathbf{h}_{k-1}+\bar{\mathbf{B}}x_k,\quad
y_k=\mathbf{C}\mathbf{h}_k.
\label{eq:ssm_discrete}
\end{equation}

\begin{figure}[t]
\centering
\includegraphics[width=.96\textwidth]{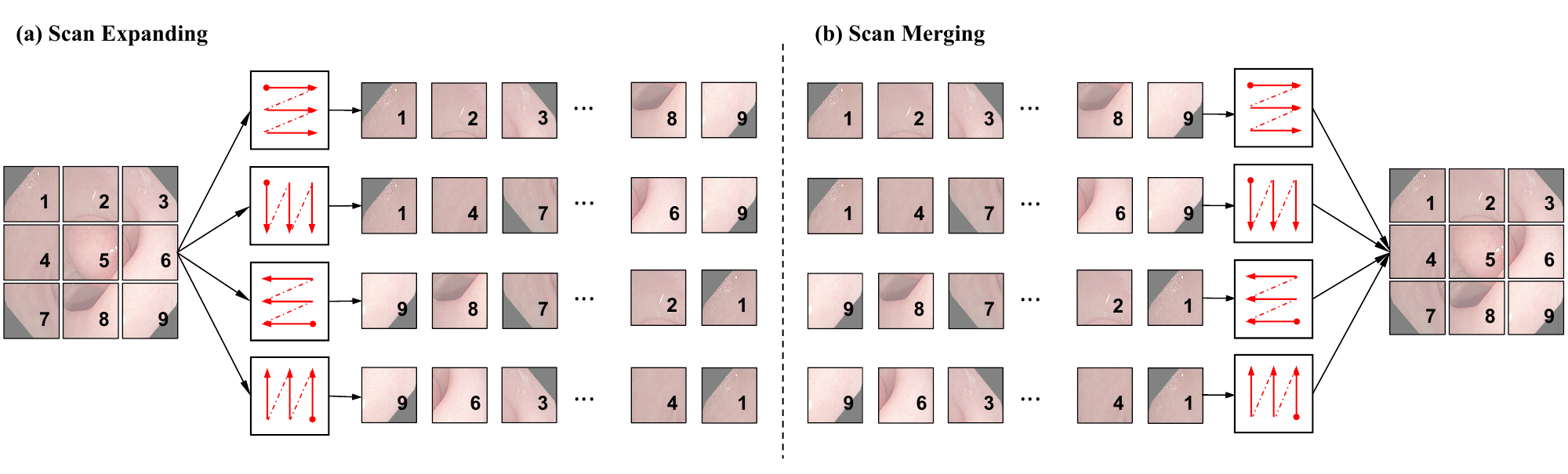}
\caption{Illustration of SS2D operations: (a) scan expanding unfolds 2D features into directional 1D scanning sequences; (b) scan merging restores and fuses the scanned sequences into a 2D feature representation.}
\label{fig:ss2d}
\end{figure}

The selective state-space mechanism in Mamba makes some parameters dependent on input tokens, enabling the model to adaptively retain or discard information according to content~\cite{gu2023mamba}. In visual tasks, the input is no longer a 1D sentence sequence but a 2D feature map. VMamba~\cite{liu2024vmamba} adopts SS2D to scan features along multiple directions, thereby approximating 2D contextual propagation. For a feature map $\mathbf{F}\in\mathbb{R}^{H'\times W'\times C}$, SS2D can be abstracted as
\begin{equation}
\operatorname{SS2D}(\mathbf{F})=
\operatorname{Merge}\left(
\mathcal{S}_{\rightarrow}(\mathbf{F}),
\mathcal{S}_{\leftarrow}(\mathbf{F}),
\mathcal{S}_{\downarrow}(\mathbf{F}),
\mathcal{S}_{\uparrow}(\mathbf{F})
\right),
\label{eq:ss2d}
\end{equation}
where $\mathcal{S}$ denotes 1D selective scanning and Merge fuses multi-directional outputs. SS2D provides global modeling with linear complexity, but it converts local 2D adjacency into scan-order relationships. As shown in Fig.~\ref{fig:ss2d}, adjacent boundary pixels may be separated after unfolding, which can weaken local continuity.

\subsection{Convolutional Scoring Gating Module}

The objective of CSG is to inject local boundary priors into state-space features. As shown in Fig.~\ref{fig:csg_vss} (a), CSG preserves the original bottleneck feature and uses a spatial scoring branch to generate a multiplicative gate. Let the input bottleneck feature be $\mathbf{F}\in\mathbb{R}^{C\times H'\times W'}$. CSG first performs channel mixing with a bias-free $1\times1$ pointwise convolution and then extracts local context using a bias-free depthwise convolution. In the main model, $k=7$. After BatchNorm and a sigmoid activation, the spatial scoring map is
\begin{equation}
\mathbf{G}=\sigma\left(\operatorname{BN}\left(
\operatorname{DWConv}_{7\times7}\left(
\operatorname{PWConv}_{1\times1}(\mathbf{F})
\right)\right)\right).
\label{eq:csg_gate}
\end{equation}
The final output is
\begin{equation}
\mathbf{F}_{\text{out}}=\mathbf{F}\odot \mathbf{G},
\label{eq:csg_output}
\end{equation}
where $\odot$ denotes element-wise multiplication. Compared with additive residual fusion, multiplicative gating provides an explicit mechanism for suppressing unreliable background responses while preserving locally salient boundary or lesion features. The $7\times7$ depthwise convolution provides a wider local field than a $3\times3$ convolution at low cost. In the controlled ablation variants, pointwise and depthwise convolutions are kept in the same order, while the kernel size ($3\times3$ or $7\times7$) and fusion rule (additive residual fusion or multiplicative gating) are varied independently. This factorial design separates the effect of the local receptive field from the effect of the gating formulation under matched training settings.

\subsection{Training Objective and Zero-Inference-Cost Training Strategies}

We optimize binary polyp segmentation using a joint BCE and Dice objective. Given the predicted probability map $\hat{\mathbf{Y}}$ and the ground-truth mask $\mathbf{Y}$, the segmentation objective for a single output branch is defined as
\begin{equation}
\mathcal{L}_{\text{seg}}=
\mathcal{L}_{\text{BCE}}(\hat{\mathbf{Y}},\mathbf{Y})
+\mathcal{L}_{\text{Dice}}(\hat{\mathbf{Y}},\mathbf{Y}),
\label{eq:loss_seg}
\end{equation}
where the Dice term directly optimizes foreground overlap and is written as
\begin{equation}
\mathcal{L}_{\text{Dice}}
=1-\frac{2\sum_i \hat{y}_i y_i+\epsilon}
{\sum_i \hat{y}_i+\sum_i y_i+\epsilon}.
\label{eq:loss_dice}
\end{equation}

To improve training stability, auxiliary heads are attached to the $16\times16$ and $32\times32$ decoder outputs. These heads project 384 and 192 channels to one channel, respectively, and upsample the predictions to $256\times256$. Let the final output be $\hat{\mathbf{Y}}_0$ and the auxiliary outputs be $\hat{\mathbf{Y}}_1$ and $\hat{\mathbf{Y}}_2$. The total objective is
\begin{equation}
\mathcal{L}_{\text{total}}
=\mathcal{L}_{\text{seg}}(\hat{\mathbf{Y}}_0,\mathbf{Y})
+0.5\mathcal{L}_{\text{seg}}(\hat{\mathbf{Y}}_1,\mathbf{Y})
+0.25\mathcal{L}_{\text{seg}}(\hat{\mathbf{Y}}_2,\mathbf{Y}).
\label{eq:loss_total}
\end{equation}

During training, EMA weights are maintained with decay 0.999. During inference, EMA weights are used and all auxiliary heads are removed. Thus, deep supervision and EMA affect only optimization and introduce no additional inference cost. This training protocol is applied consistently to VM-UNet, VM-UNetV2, and CSG-Mamba; therefore, the VM-UNet control baseline differs from CSG-Mamba by removing the proposed CSG module while keeping the optimization setting matched.

\section{Experimental Setup}
\subsection{Datasets}

Kvasir-SEG~\cite{jha2020kvasir} was used for training and in-domain evaluation, with a fixed split of 800 training, 100 validation, and 100 test images. For zero-shot cross-domain evaluation, the model trained on Kvasir-SEG was directly tested on all 612 CVC-ClinicDB~\cite{bernal2015clinicdb} images and all 380 CVC-ColonDB~\cite{tajbakhsh2016colondb} images, without fine-tuning or adaptation.

\subsection{Implementation Details}

All images were resized to $256\times256$. Training augmentations included random horizontal and vertical flipping with probability 0.5 and random rotation with probability 0.5, with rotation angles sampled from $[0^\circ, 360^\circ]$. Validation and test images were resized and normalized using the Kvasir mean and standard deviation, followed by per-image min-max stretching to $[0,255]$.

Models were trained for 300 epochs with a batch size of 32. AdamW~\cite{loshchilov2019adamw} used a learning rate of $1\times10^{-4}$, $\beta=(0.9,0.999)$, $\epsilon=10^{-8}$, and a weight decay of $10^{-2}$. Cosine annealing~\cite{loshchilov2017sgdr} used $T_{\max}=300$ and $\eta_{\min}=10^{-5}$ without warmup. The prediction threshold was fixed at 0.5. Validation was performed at epoch 1 and every 30 epochs, and the checkpoint with the highest validation Dice was selected.

For a fair comparison among Vision Mamba variants, VM-UNet, VM-UNetV2, and CSG-Mamba all used DropPath 0.3, EMA 0.999, and the same deep supervision setting. All core results were obtained with three random seeds (42, 123, and 999) and are reported as the mean $\pm$ standard deviation. Metrics include mIoU, Dice, HD95, Accuracy, and Sensitivity, covering region overlap, boundary error, and lesion recall~\cite{maierhein2024metrics,taha2015metrics}.

\subsection{Compared Methods}

We compared CSG-Mamba with U-Net~\cite{ronneberger2015unet}, ResUNet~\cite{zhang2018resunet}, PraNet~\cite{fan2020pranet}, TransUNet~\cite{chen2021transunet}, VM-UNet~\cite{ruan2024vmunet}, and VM-UNetV2~\cite{zhang2024vmunetv2}. All methods used the same split, input size, and evaluation protocol. CSG-Mamba uses the VMamba-Tiny pretrained backbone, a mirror-style VM-UNet decoder, and the proposed CSG module. VM-UNet is the strict structural control baseline without CSG but with the same deep supervision, EMA, and DropPath settings, while VM-UNetV2 is treated as an independent Mamba-based comparison method trained under the same stabilization protocol.

\section{Experimental Results and Analysis}
\subsection{Main Results}

\begin{figure}[t]
\centering
\includegraphics[width=.98\textwidth]{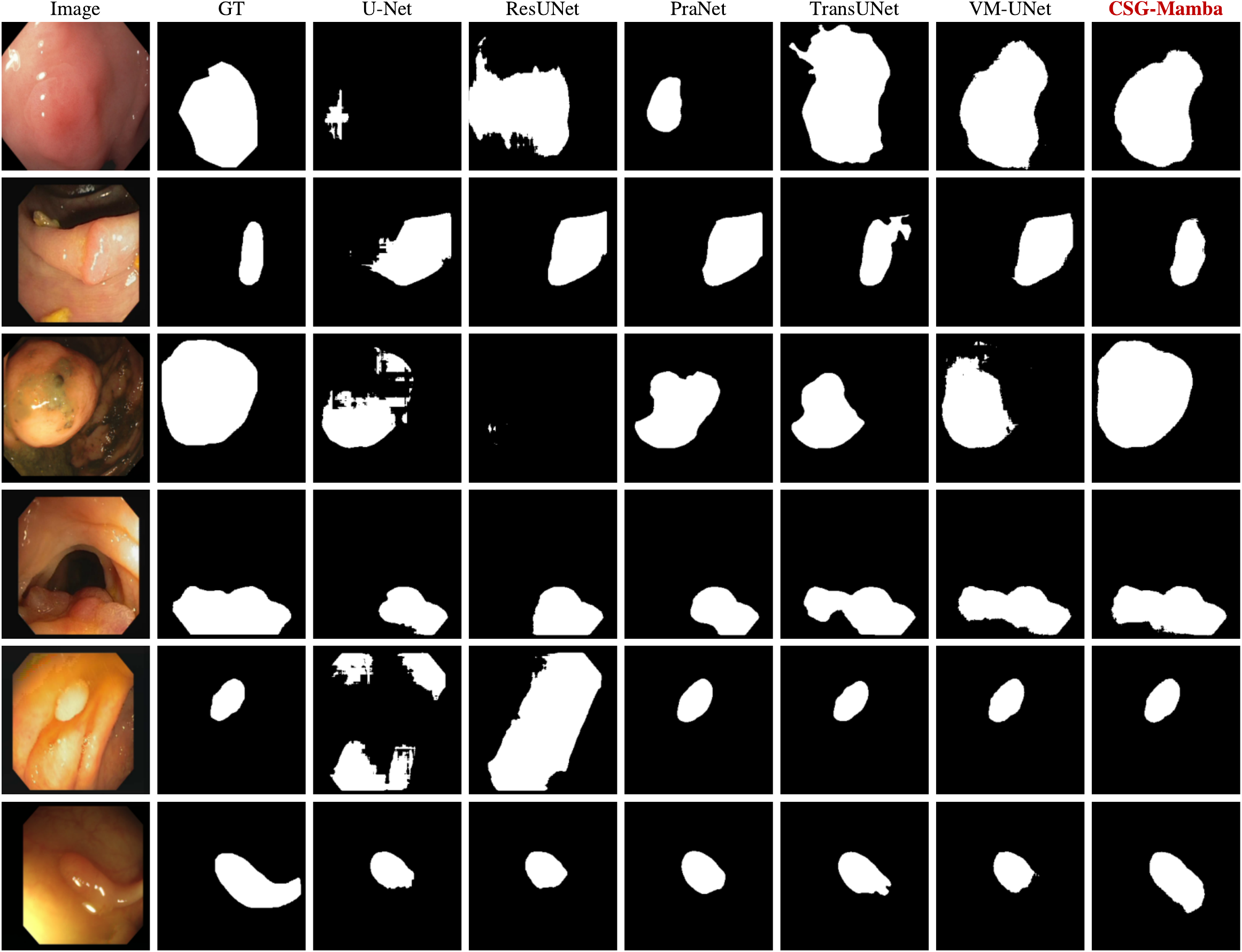}
\caption{Qualitative segmentation results. Columns show the image, ground truth, U-Net, ResUNet, PraNet, TransUNet, VM-UNet, and CSG-Mamba.}
\label{fig:qualitative}
\end{figure}

Fig.~\ref{fig:qualitative} shows qualitative comparisons on challenging samples. CNN models may produce contour shrinkage or boundary discontinuities, Transformer models may introduce false positives around highlights or folds, and VM-UNet can over-smooth boundaries. In these examples, CSG-Mamba more closely follows fragmented edges and suppresses lesion-like background responses.

Table~\ref{tab:kvasir} reports in-domain results on the Kvasir-SEG test set. CSG-Mamba achieves the best mIoU, Dice, HD95, Accuracy, and Sensitivity. Compared with VM-UNet, it improves Dice by 0.0042 and reduces HD95 by 2.38 with a moderate parameter increase. Compared with TransUNet, it uses approximately one-third of the parameters while obtaining a higher Dice score.

\begin{table}[t]
\caption{In-domain evaluation results on Kvasir-SEG (mean$\pm$std).}
\label{tab:kvasir}
\centering
\scriptsize
\setlength{\tabcolsep}{3pt}
\resizebox{\textwidth}{!}{%
\begin{tabular}{lcccccc}
\toprule
Model & mIoU & Dice & HD95$\downarrow$ & Accuracy & Sensitivity & Params \\
\midrule
U-Net & $0.7875\pm0.0017$ & $0.8572\pm0.0006$ & $33.30\pm1.49$ & $0.9687\pm0.0010$ & $0.8804\pm0.0068$ & 31.04M \\
ResUNet & $0.7995\pm0.0094$ & $0.8656\pm0.0081$ & $30.67\pm3.98$ & $0.9703\pm0.0010$ & $0.8878\pm0.0053$ & 63.85M \\
PraNet & $0.8453\pm0.0027$ & $0.9006\pm0.0014$ & $20.66\pm1.57$ & $0.9764\pm0.0017$ & $0.9169\pm0.0117$ & 24.51M \\
TransUNet & $0.8540\pm0.0064$ & $0.9082\pm0.0061$ & $17.69\pm0.71$ & $0.9776\pm0.0005$ & $0.9184\pm0.0094$ & 99.04M \\
VM-UNet & $0.8620\pm0.0034$ & $0.9178\pm0.0020$ & $18.25\pm1.42$ & $0.9804\pm0.0009$ & $0.9284\pm0.0042$ & 27.43M \\
VM-UNetV2 & $0.8661\pm0.0021$ & $0.9180\pm0.0016$ & $17.40\pm1.76$ & $0.9809\pm0.0010$ & $0.9278\pm0.0095$ & 27.63M \\
\textbf{CSG-Mamba} & $\mathbf{0.8677\pm0.0049}$ & $\mathbf{0.9220\pm0.0031}$ & $\mathbf{15.87\pm0.48}$ & $\mathbf{0.9817\pm0.0008}$ & $\mathbf{0.9301\pm0.0017}$ & 32.30M \\
\bottomrule
\end{tabular}}
\end{table}

Table~\ref{tab:clinicdb} reports zero-shot results on the full CVC-ClinicDB test set. CSG-Mamba achieves the best mIoU, Dice, and Sensitivity, while VM-UNet obtains the lowest HD95 and VM-UNetV2 has slightly higher Accuracy. Compared with VM-UNet, CSG-Mamba improves mIoU, Dice, and Sensitivity by 0.0076, 0.0056, and 0.0125, respectively, with a 0.90 increase in HD95.

\begin{table}[t]
\caption{Zero-shot test results on the full CVC-ClinicDB dataset (mean$\pm$std).}
\label{tab:clinicdb}
\centering
\scriptsize
\setlength{\tabcolsep}{3pt}
\resizebox{\textwidth}{!}{%
\begin{tabular}{lccccc}
\toprule
Model & mIoU & Dice & HD95$\downarrow$ & Accuracy & Sensitivity \\
\midrule
U-Net & $0.5922\pm0.0093$ & $0.6762\pm0.0111$ & $73.10\pm4.79$ & $0.9539\pm0.0010$ & $0.7077\pm0.0238$ \\
ResUNet & $0.5760\pm0.0278$ & $0.6602\pm0.0264$ & $75.48\pm4.84$ & $0.9451\pm0.0070$ & $0.7199\pm0.0082$ \\
PraNet & $0.7403\pm0.0181$ & $0.8163\pm0.0167$ & $34.73\pm3.98$ & $0.9684\pm0.0032$ & $0.8449\pm0.0139$ \\
TransUNet & $0.7497\pm0.0031$ & $0.8260\pm0.0017$ & $30.27\pm0.80$ & $0.9697\pm0.0008$ & $0.8434\pm0.0065$ \\
VM-UNet & $0.7689\pm0.0050$ & $0.8437\pm0.0044$ & $\mathbf{24.72\pm1.02}$ & $0.9751\pm0.0015$ & $0.8489\pm0.0092$ \\
VM-UNetV2 & $0.7643\pm0.0057$ & $0.8364\pm0.0066$ & $30.04\pm2.48$ & $\mathbf{0.9754\pm0.0009}$ & $0.8363\pm0.0150$ \\
\textbf{CSG-Mamba} & $\mathbf{0.7765\pm0.0079}$ & $\mathbf{0.8493\pm0.0074}$ & $25.62\pm1.28$ & $0.9753\pm0.0008$ & $\mathbf{0.8614\pm0.0069}$ \\
\bottomrule
\end{tabular}}
\end{table}

Table~\ref{tab:colondb} reports zero-shot results on the more challenging CVC-ColonDB test set. CSG-Mamba achieves the best mIoU, Dice, Accuracy, and Sensitivity, improving Dice over VM-UNet and VM-UNetV2 by 0.0133 and 0.0205, respectively. VM-UNet has slightly lower HD95, but CSG-Mamba reduces HD95 by 14.70 compared with VM-UNetV2.

\begin{table}[t]
\caption{Zero-shot test results on CVC-ColonDB (mean$\pm$std).}
\label{tab:colondb}
\centering
\scriptsize
\setlength{\tabcolsep}{3pt}
\resizebox{\textwidth}{!}{%
\begin{tabular}{lccccc}
\toprule
Model & mIoU & Dice & HD95$\downarrow$ & Accuracy & Sensitivity \\
\midrule
U-Net & $0.4883\pm0.0281$ & $0.5603\pm0.0294$ & $109.21\pm11.37$ & $0.9519\pm0.0029$ & $0.5677\pm0.0241$ \\
ResUNet & $0.4597\pm0.0183$ & $0.5270\pm0.0208$ & $123.58\pm9.22$ & $0.9512\pm0.0037$ & $0.5359\pm0.0205$ \\
PraNet & $0.6029\pm0.0179$ & $0.6811\pm0.0215$ & $69.77\pm8.07$ & $0.9632\pm0.0020$ & $0.6829\pm0.0265$ \\
TransUNet & $0.5878\pm0.0171$ & $0.6670\pm0.0196$ & $53.42\pm10.22$ & $0.9535\pm0.0015$ & $0.6947\pm0.0252$ \\
VM-UNet & $0.6399\pm0.0068$ & $0.7285\pm0.0060$ & $\mathbf{39.70\pm0.74}$ & $0.9629\pm0.0004$ & $0.7372\pm0.0101$ \\
VM-UNetV2 & $0.6436\pm0.0082$ & $0.7213\pm0.0100$ & $56.65\pm5.10$ & $0.9662\pm0.0004$ & $0.7233\pm0.0055$ \\
\textbf{CSG-Mamba} & $\mathbf{0.6570\pm0.0097}$ & $\mathbf{0.7418\pm0.0091}$ & $41.95\pm3.47$ & $\mathbf{0.9676\pm0.0018}$ & $\mathbf{0.7519\pm0.0063}$ \\
\bottomrule
\end{tabular}}
\end{table}

\begin{figure}[t]
\centering
\includegraphics[width=.86\textwidth]{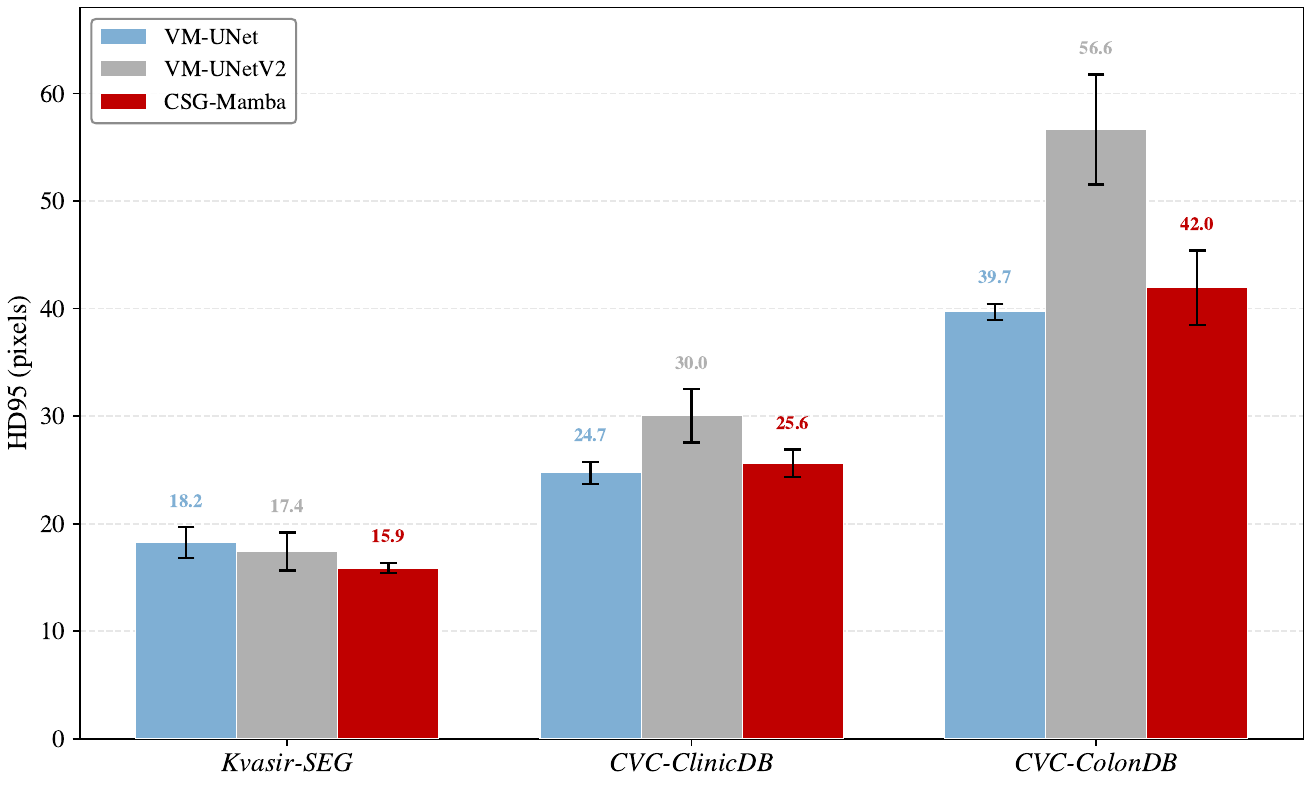}
\caption{Cross-domain HD95 comparison with error bars over three random seeds.}
\label{fig:hd95}
\end{figure}

Fig.~\ref{fig:hd95} compares HD95 across datasets. CSG-Mamba obtains the lowest HD95 on Kvasir-SEG and substantially mitigates the boundary degradation of VM-UNetV2 on the CVC-ColonDB test set. On CVC-ClinicDB, its HD95 is slightly higher than that of VM-UNet but remains close.

\subsection{Parameter Efficiency Analysis}

\begin{figure}[t]
\centering
\includegraphics[width=.86\textwidth]{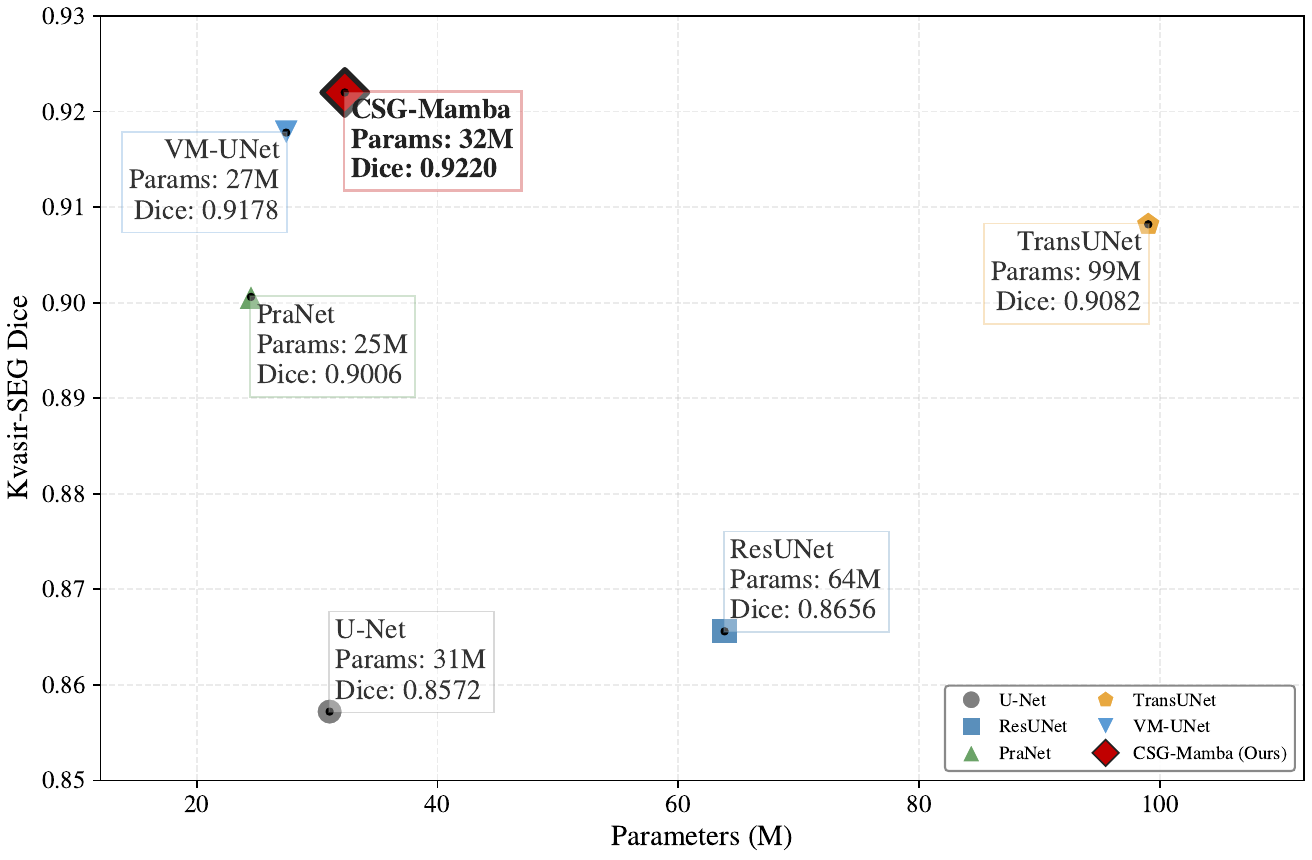}
\caption{Parameter efficiency scatter plot on Kvasir-SEG.}
\label{fig:efficiency}
\end{figure}

Fig.~\ref{fig:efficiency} shows parameter efficiency. CSG-Mamba has a parameter scale similar to U-Net but improves Dice by 0.0648. It is also substantially smaller than TransUNet while achieving a higher Dice score, suggesting that the additional CSG parameters are effective for boundary modeling.

\subsection{Ablation on the CSG Mechanism}

Table~\ref{tab:ablation_csg} reports a controlled mechanism ablation at the bottleneck. To avoid conflating several design choices, this experiment uses a $2\times2$ factorial design to decouple two factors: the depthwise convolution kernel size ($3\times3$ or $7\times7$) and the feature fusion rule (additive residual fusion with ReLU or multiplicative gating with sigmoid). All variants are evaluated with the same three-seed protocol and the same stabilization settings as the main Mamba-based experiments. This design evaluates two separate questions: whether a wider local field is useful and whether multiplicative gating changes boundary behavior beyond merely adding a convolutional branch.

\begin{table}[t]
\caption{Ablation on the CSG mechanism at the bottleneck (mean$\pm$std).}
\label{tab:ablation_csg}
\centering
\footnotesize
\setlength{\tabcolsep}{1.2pt}
\resizebox{\textwidth}{!}{%
\begin{tabular}{@{}lccccccc@{}}
\toprule
\shortstack{Bottleneck\\variant} & \shortstack{Kvasir\\HD95$\downarrow$} & \shortstack{Kvasir\\Dice} & \shortstack{ClinicDB\\HD95$\downarrow$} & \shortstack{ClinicDB\\Dice} & \shortstack{ColonDB\\HD95$\downarrow$} & \shortstack{ColonDB\\Dice} & Params \\
\midrule
None & $18.25\pm1.42$ & $0.9178\pm0.0020$ & $\mathbf{24.72\pm1.02}$ & $0.8437\pm0.0044$ & $\mathbf{39.70\pm0.74}$ & $0.7285\pm0.0060$ & 27.43M \\
DW $3\times3$ + Add & $17.41\pm0.36$ & $0.9240\pm0.0035$ & $26.45\pm1.47$ & $0.8482\pm0.0002$ & $41.40\pm4.98$ & $\mathbf{0.7441\pm0.0019}$ & 32.18M \\
DW $3\times3$ + Gate & $15.70\pm1.29$ & $0.9237\pm0.0031$ & $27.28\pm2.31$ & $0.8465\pm0.0061$ & $46.65\pm7.02$ & $0.7348\pm0.0083$ & 32.18M \\
DW $7\times7$ + Add & $\mathbf{15.59\pm0.84}$ & $\mathbf{0.9260\pm0.0033}$ & $27.15\pm2.03$ & $0.8447\pm0.0038$ & $46.00\pm1.12$ & $0.7292\pm0.0020$ & 32.30M \\
\textbf{DW $7\times7$ + Gate (CSG)} & $15.87\pm0.48$ & $0.9220\pm0.0031$ & $25.62\pm1.28$ & $\mathbf{0.8493\pm0.0074}$ & $41.95\pm3.47$ & $0.7418\pm0.0091$ & 32.30M \\
\bottomrule
\end{tabular}}
\end{table}

Here, Add denotes additive residual fusion with ReLU, and Gate denotes sigmoid-based multiplicative gating. Because Kvasir Dice is concentrated in a narrow high-performance range, HD95 provides a more sensitive indication of boundary behavior. With a fixed $3\times3$ kernel, replacing additive fusion with gating reduces Kvasir HD95 from 17.41 to 15.70 pixels while keeping Dice nearly unchanged, indicating that gating mainly affects localization rather than regional overlap.

The $7\times7$ additive branch gives the best in-domain Dice and HD95, showing the benefit of a wider local field, but it increases cross-domain HD95 on ClinicDB and ColonDB. With the same $7\times7$ kernel, multiplicative gating reduces these HD95 values to 25.62 and 41.95 and improves the external Dice values to 0.8493 and 0.7418. Thus, CSG is selected for a balanced boundary-oriented trade-off: large-kernel convolution supplies local context, while multiplicative gating regulates this context under domain shift.

\subsection{Ablation on CSG Placement}

Table~\ref{tab:ablation_placement} compares CSG insertion positions under the same three-seed protocol, with VM-UNet (``none'') as the reference. Bottleneck-only achieves the best Kvasir HD95 (15.87), the best ClinicDB Dice (0.8493), and a strong ColonDB Dice (0.7418). Decoder-only gives the highest Kvasir Dice but sharply increases ColonDB HD95, suggesting sensitivity to in-domain texture patterns. Encoder-only obtains the best ColonDB Dice at the cost of weaker Kvasir and ClinicDB overlap, while insertion at all positions adds parameters and degrades Kvasir HD95. Overall, bottleneck placement offers the most balanced trade-off between in-domain boundary quality, cross-domain generalization, and parameter cost.

\begin{table}[t]
\caption{Ablation on CSG placement (mean$\pm$std).}
\label{tab:ablation_placement}
\centering
\footnotesize
\setlength{\tabcolsep}{1.2pt}
\resizebox{\textwidth}{!}{%
\begin{tabular}{@{}lccccccc@{}}
\toprule
Position & \shortstack{Kvasir\\HD95$\downarrow$} & \shortstack{Kvasir\\Dice$\uparrow$} & \shortstack{ClinicDB\\HD95$\downarrow$} & \shortstack{ClinicDB\\Dice$\uparrow$} & \shortstack{ColonDB\\HD95$\downarrow$} & \shortstack{ColonDB\\Dice$\uparrow$} & Params \\
\midrule
none & $18.25\pm1.42$ & $0.9178\pm0.0020$ & $\mathbf{24.72\pm1.02}$ & $0.8437\pm0.0044$ & $\mathbf{39.70\pm0.74}$ & $0.7285\pm0.0060$ & 27.43M \\
encoder only & $17.82\pm1.83$ & $0.9155\pm0.0051$ & $26.25\pm0.84$ & $0.8398\pm0.0023$ & $42.37\pm6.27$ & $\mathbf{0.7431\pm0.0169}$ & 29.11M \\
decoder only & $16.96\pm1.26$ & $\mathbf{0.9235\pm0.0042}$ & $29.91\pm0.87$ & $0.8389\pm0.0009$ & $58.09\pm13.13$ & $0.7147\pm0.0256$ & 33.94M \\
\textbf{bottleneck only} & $\mathbf{15.87\pm0.48}$ & $0.9220\pm0.0031$ & $25.62\pm1.28$ & $\mathbf{0.8493\pm0.0074}$ & $41.95\pm3.47$ & $0.7418\pm0.0091$ & 32.30M \\
all & $19.31\pm2.77$ & $0.9145\pm0.0129$ & $27.80\pm1.17$ & $0.8418\pm0.0067$ & $49.89\pm6.80$ & $0.7353\pm0.0101$ & 40.50M \\
\bottomrule
\end{tabular}}
\end{table}

\section{Discussion}

The results suggest that Vision State Space Models benefit from cooperation with local inductive biases rather than relying solely on sequential state-space scanning. State-space scanning provides efficient long-range modeling, but it is not naturally aligned with pixel-level boundary prediction. The mechanism ablation shows that the large depthwise kernel supplies local boundary context, while multiplicative gating improves the cross-domain HD95 trade-off under the $7\times7$ setting without merely amplifying domain-specific textures.

The ClinicDB results also show that cross-domain generalization cannot be summarized by a single metric. CSG-Mamba obtains the highest mIoU, Dice, and Sensitivity on ClinicDB, although VM-UNet has a slightly lower HD95. On ColonDB, CSG-Mamba obtains stronger region overlap and recall, while VM-UNet remains slightly better in HD95. This pattern suggests that CSG tends to retain suspected lesion regions, which may reduce missed detections but can also lead to mild boundary expansion.

Several limitations remain. The experiments focus on 2D polyp segmentation, and the generalizability of CSG to other medical tasks requires further study. Although the Mamba-based comparisons share the same deep supervision, EMA, and DropPath settings, future work can further evaluate CSG under additional training recipes and on data from more clinical centers.

\section{Conclusion}

This paper proposes CSG-Mamba for endoscopic polyp segmentation. To compensate for the weakened 2D continuity caused by SS2D sequence unfolding, CSG-Mamba inserts a lightweight convolutional scoring gating module at the bottleneck. Under a matched zero-inference-cost training protocol for the Mamba-based methods, CSG-Mamba achieves the best Dice and HD95 on Kvasir-SEG, the best mIoU, Dice, Accuracy, and Sensitivity on CVC-ColonDB, and the best mIoU, Dice, and Sensitivity on CVC-ClinicDB. These results indicate that combining local convolutional priors with global state-space modeling can improve boundary quality and cross-domain robustness.

\begin{credits}
\subsubsection{\discintname}
The authors have no competing interests to declare that are relevant to the content of this article.
\end{credits}

\bibliographystyle{splncs04}
\bibliography{refs}
\end{document}